# Qualitative Possibilistic Mixed-Observable MDPs

**Nicolas Drougard, Jean-Loup Farges, Florent Teichteil-Königsbuch**
Onera – The French Aerospace Lab
2 avenue Edouard Belin
31055 Toulouse Cedex 4, France

**Didier Dubois**
IRIT – Paul Sabatier University
118 route de Narbonne
31062 Toulouse Cedex 4, France

## Abstract

Possibilistic and qualitative POMDPs ($\pi$-POMDPs) are counterparts of POMDPs used to model situations where the agent's initial belief or observation probabilities are imprecise due to lack of past experiences or insufficient data collection. However, like probabilistic POMDPs, optimally solving $\pi$-POMDPs is intractable: the finite belief state space exponentially grows with the number of system's states. In this paper, a possibilistic version of Mixed-Observable MDPs is presented to get around this issue: the complexity of solving $\pi$-POMDPs, some state variables of which are fully observable, can be then dramatically reduced. A value iteration algorithm for this new formulation under infinite horizon is next proposed and the optimality of the returned policy (for a specified criterion) is shown assuming the existence of a "stay" action in some goal states. Experimental work finally shows that this possibilistic model outperforms probabilistic POMDPs commonly used in robotics, for a target recognition problem where the agent's observations are imprecise.

## 1 INTRODUCTION

Markov Decision Processes (MDPs) define a useful formalism to express sequential decision problems under uncertainty [2]. Partially Observable MDPs (POMDPs) [15] are used to model situations in which an agent does not know directly the current state of the system: its decisions are based on a probability distribution over the state space. This distribution known as "belief" is updated at each stage $t \in \mathbb{N}$ of the process using the current observation. This update based on Bayes' rule needs perfect knowledge of the agent's initial belief and of the transition and observation probability distributions.

Consider situations where the agent totally ignores the system's initial state, for instance a robot that is for the first time in a room with an unknown exit location (initial belief) and has to find the exit and reach it. In practice, no experience can be repeated in order to extract a frequency of the exit's location. In this kind of situation, uncertainty is not due to a random fact, but to a lack of knowledge: no frequentist initial belief can be used to define the model. A uniform probability distribution is often chosen in order to assign the same mass to each state. This choice can be justified based on the subjective probability theory [5] (the probability distribution represents then an exchangeable bet) but subjective probabilities and observation frequencies are combined during the belief update.

In other cases, the agent may strongly believe that the exit is located in a wall as in the vast majority of rooms, but it still grants a very small probability $p_\epsilon$ to the fact that the exit may be a staircase in the middle of the room. Even if this is very unlikely to be the case, this second option must be taken into account in the belief, otherwise Bayes' rule cannot correctly update it if the exit is actually in the middle of the room. Eliciting $p_\epsilon$ without past experience is not obvious at all and does not rely on any rational reasons, yet it dramatically impacts the agent's policy. On the contrary, possibilistic uncertainty models allow the agent to elicit beliefs with imprecise unbiased knowledge.

The $\pi$-POMDP model is a possibilistic and qualitative counterpart of the probabilistic POMDP model [12]: it allows a formal modeling of total ignorance using a possibility distribution equal to 1 on all the states. This distribution means that all states are equally possible independently of how likely they are to happen (no necessary state). Moreover, consider robotic missions using visual perception: observations of the robot agent are outputs of image processing algorithms whose semantics (image correlation, object matching, class inference, etc.) is so complex that probabilities of occurrence are hard to rigorously extract. Finding qualitative estimates of their recognition performance is easier: the $\pi$-POMDP model only require qualitative data, thus it allows to construct the model without using more information than really available.

However, just like the probabilistic POMDP model, this possibilistic model faces the difficulty of computing an optimal policy. Indeed, the size of its belief space exponentially grows with the size of the state space, which prevents the use of $\pi$-POMDPs in practice. In situations where most state variables are fully observable, an alternative structuring of the model still allows to solve the problem: the possibilistic Mixed-Observable MDP model ($\pi$-MOMDP), which is the first contribution of this paper, indeed allows us to reason with beliefs over the partially observed states only. In this model borrowed from probabilistic POMDPs [9, 1], states are factorized into two sets of fully and partially observable state variables. Whereas, in probabilistic POMDPs, this factorized model permits to reason about smaller *continuous* belief subspaces and speed up $\alpha$-vector operations, its impact is totally different in possibilistic POMDPs: it allows us to reduce the size of the *discrete* belief state space.

Our second contribution is a possibilistic value iteration algorithm for this extension, which exploits the hybrid structure of the belief space. This algorithm is derived from a $\pi$-MDP algorithm, Sabbadin's work update, whose optimality of the returned policy is proved for an infinite horizon criterion which is made explicit. It relies on intermediate "stay" actions that are needed to guarantee convergence of the algorithm but that vanish in the optimized policy for non goal states; they are the possibilistic counterparts of the discount factor in probabilistic POMDPs.

Finally, we experimentally demonstrate that in some situations $\pi$-MOMDPs outperform their probabilistic counterparts for instance on information-gathering robotic problems where the observation function resulting from complex image processing algorithms is not precisely known, as it is often the case in realistic applications. This result is significant for us, because roboticists commonly think that probabilistic POMDPs, and more generally Bayesian approaches, are first-choice reasoning models to solve sequential information-gathering missions. We prove in this paper that sometimes possibilistic uncertainty models perform better in practice.

## 2 BACKGROUND

The Markov Decision Process framework models situations in which the *system*, for instance the physical part of an agent and its environment, has a Markovian dynamic over time. The different possible states of the system are represented by the elements $s$ of the finite state space $\mathcal{S}$. The initial system state is denoted by $s_0 \in \mathcal{S}$. At each stage of the process, modeled by non negative integers $t \in \mathbb{N}$, the decisional part of the agent can choose an *action* $a$ in the finite set $\mathcal{A}$. The chosen action $a_t$ determines the uncertainty over the future state $s_{t+1}$ knowing the current state $s_t$.

### 2.1 Qualitative possibilistic MDPs

The work of Sabbadin [12] proposes a possibilistic counterpart of Markov Decision Processes. In this framework, the transition uncertainty is modeled as *qualitative possibility distributions* over $\mathcal{S}$. Let $\mathcal{L}$ be the *possibility scale i.e.* a finite and totally ordered set whose greatest element is denoted by $1_\mathcal{L}$ and the least element by $0_\mathcal{L}$ (classically $\mathcal{L}=\{0, \frac{1}{k}, \frac{2}{k}, \ldots, \frac{k-1}{k}, 1\}$ with $k \in \mathbb{N}^*$). A qualitative possibility distribution over $\mathcal{S}$ is a function $\pi : \mathcal{S} \to \mathcal{L}$ such that $\max_s \pi(s) = 1_\mathcal{L}$ (possibilistic normalization), implying that at least one entirely possible state exists. Inequality $\pi(s) < \pi(s')$ means that state $s'$ is more plausible than state $s$. This modeling needs less information than the probabilistic one since the plausibilities of events are "only" classified (in $\mathcal{L}$) but not quantified.

The transition function $T^\pi$ is defined as follows: for each pair of states $(s, s') \in \mathcal{S}^2$ and action $a \in \mathcal{A}$, $T^\pi(s, a, s') = \pi(s' \mid s, a) \in \mathcal{L}$, the possibility of reaching the system state $s'$ conditionned on the current state $s$ and action $a$. Scale $\mathcal{L}$ serves as well to model the *preference* over states: function $\mu : \mathcal{S} \to \mathcal{L}$ models the agent's preferences. A $\pi$-MDP is then entirely defined by the tuple $\langle \mathcal{S}, \mathcal{A}, \mathcal{L}, T^\pi, \mu \rangle$.

A *policy* is a sequence $(\delta_t)_{t \geqslant 0}$ of *decision rules* $\delta : \mathcal{S} \to \mathcal{A}$ indexed by the stage of the process $t \in \mathbb{N}$: $\delta_t(s)$ is the action executed in state $s$ at decision epoch $t$. We denote by $\Delta_p$ the set of all $p$-sized policies $(\delta_0, \ldots, \delta_{p-1})$. Let $\tau = (s_1, \ldots, s_p) \in \mathcal{S}^p$ be a $p$-sized trajectory and $(\delta) = (\delta_t)_{t=0}^{p-1}$ a $p$-sized policy. The set of all the $p$-sized trajectories is denoted by $\mathcal{T}_p$.

The sequence $(s_t)_{t \geqslant 0}$ is a Markov process: the possibility of the trajectory $\tau = (s_1, \ldots, s_p)$ which starts from $s_0$ using $(\delta) \in \Delta_p$ is then

$$\Pi(\tau \mid s_0, (\delta)) = \min_{t=0}^{p-1} \pi(s_{t+1} \mid s_t, \delta_t(s_t)).$$

We define the preference of $\tau \in \mathcal{T}_p$ as the preference of the last state: $M(\tau) = \mu(s_p)$. As advised in [13] for problems in which there is no risk of being blocked in an unsatisfactory state, we use here the optimistic qualitative decision criterion [3] which is the Sugeno integral of the preference distribution over trajectories using possibility measure:

$$u_p(s_0, (\delta)) = \max_{\tau \in \mathcal{T}_p} \min \{\Pi(\tau \mid s_0, (\delta)), M(\tau)\}. \quad (1)$$

A policy which maximizes criterion 1 ensures that there exists a possible and satisfactory $p$-sized trajectory. The finite horizon $\pi$-MDP is solved when such a policy is found. The optimal $p$-sized horizon criterion $u_p^*(s) = \max_{(\delta) \in \Delta_p} u_p(s, (\delta))$ is the solution of the following dynamic programming equation, as proved in [6]: $\forall i \in \{1 \ldots p\}, \forall s \in \mathcal{S}$,

$$u_i^*(s) = \max_{a \in \mathcal{A}} \max_{s' \in \mathcal{S}} \min \left\{ \pi\left(s' \mid s, a\right), u_{i-1}^*(s') \right\}, \quad (2)$$

$$\delta_{p-i}^*(s) \in \operatorname*{argmax}_{a \in \mathcal{A}} \max_{s' \in \mathcal{S}} \min \left\{ \pi\left(s' \mid s, a\right), u_{i-1}^*(s') \right\}$$

with the initialization $u_0^*(s) = \mu(s)$.

### 2.2 The partially observable case

A possibilistic counterpart of POMDPs is also given in [12]. As in the probabilistic framework, the agent does not directly observe the system's states. Here, uncertainty over observations is also modeled as possibility distributions. The observation function $\Omega^\pi$ is defined as follows: $\forall o' \in \mathcal{O}, s \in \mathcal{S}$ and $a \in \mathcal{A}$, $\Omega^\pi(s', a, o') = \pi(o' \mid s, a)$ the possibility of the current observation $o'$ conditionned on the current state $s'$ and the previous action $a$. Then a $\pi$-POMDP is defined by the tuple $\langle \mathcal{S}, \mathcal{A}, \mathcal{L}, T^\pi, \mathcal{O}, \Omega^\pi, \mu, \beta_0 \rangle$, where $\beta_0$ is the initial possibilistic belief. The belief of the agent is a possibility distribution over $\mathcal{S}$; total ignorance is defined by a belief equal to $1_\mathcal{L}$ on all states, whereas a given state $s$ is perfectly known if the belief is equal to $1_\mathcal{L}$ on this state and to $0_\mathcal{L}$ on all other states.

The translation into $\pi$-MDP can be done in a similar way as for POMDPs: we denote by $B^\pi \subset \mathcal{L}^\mathcal{S}$ the possibilistic belief state space which contains all the possibility distributions defined on $\mathcal{S}$. We can now compute the possibilistic belief update. If at time $t$ the current belief is $\beta_t \in B^\pi$ and the agent executes action $a_t$, the belief over the future states is given by:

$$\beta_{t+1}^{a_t}(s') = \max_{s \in \mathcal{S}} \min \left\{ \pi\left(s' \mid s, a_t\right), \beta_t(s) \right\},$$

and the belief over the observations

$$\beta_{t+1}^{a_t}(o') = \max_{s' \in \mathcal{S}} \min \left\{ \pi\left(o' \mid s', a_t\right), \beta_{t+1}^{a_t}(s') \right\}.$$

Next, if the agent observes $o_{t+1} \in \mathcal{O}$, the possibilistic counterpart of Bayes' rule ensures that

$$\beta_{t+1}(s') = \begin{cases} 1_\mathcal{L} & \text{if } \beta_{t+1}^{a_t}(o_{t+1}) = \pi\left(s', o_{t+1} \mid \beta_t, a_t\right) > 0_\mathcal{L} \\ \pi\left(s', o_{t+1} \mid \beta_t, a_t\right) & \text{otherwise} \end{cases} \quad (3)$$

where $\forall (s', o') \in \mathcal{S} \times \mathcal{O}, \pi(s', o' \mid \beta, a) = \min \left\{ \pi\left(o' \mid s', a\right), \beta^a(s') \right\}$ is the joint possibility of $(s', o')$. Such an update of a belief $\beta$ is denoted by $\beta^{a, o'}$: $\beta_{t+1} = \beta_t^{a, o'}$. As the belief update is now made explicit, its dynamics can be computed: let $\Gamma^{\beta, a}(\beta') = \left\{ o' \in \mathcal{O} \mid \beta^{a, o'} = \beta' \right\}$. Then $\pi(\beta' \mid \beta, a) = \max_{o' \in \Gamma^{\beta, a}(\beta')} \beta^a(o')$ with the convention $\max_\emptyset = 0_\mathcal{L}$.

We now define the preference over belief states with a pessimistic form in order to prefer informative beliefs: a belief state has a good preference when it is unlikely that the system is in an unsatisfactory state: $\mu(\beta) = \min_{s \in \mathcal{S}} \max \left\{ \mu(s), n(\beta(s)) \right\}$, with $n : \mathcal{L} \to \mathcal{L}$ the order reversing map i.e. the only decreasing function from $\mathcal{L}$ to $\mathcal{L}$. A $\pi$-MDP is then defined, and the new dynamic programming equation is $\forall i \in \{1, \ldots, p\}, \forall \beta \in B^\pi$,

$$\begin{aligned} u_p^*(\beta) &= \max_{a \in \mathcal{A}} \max_{\beta' \in B^\pi} \min \left\{ \pi\left(\beta' \mid \beta, a\right), u_{p-1}^*(\beta') \right\} \\ &= \max_{a \in \mathcal{A}} \max_{o' \in \mathcal{O}} \min \left( \beta^a(o'), u_{p-1}^*(\beta^{a, o'}) \right) \end{aligned}$$

with the initialization $u_0^*(\beta) = \mu(\beta)$. Note that $B^\pi$ is a finite set of cardinal $\#B^\pi = \#\mathcal{L}^{\#\mathcal{S}} - (\#\mathcal{L} - 1)^{\#\mathcal{S}}$ (the total number of $\#\mathcal{S}$-size vectors valued in $\mathcal{L}$, minus $(\#\mathcal{L} - 1)^{\#\mathcal{S}}$ non-normalized distributions). However, for concrete problems, the state space can be dramatically large: $\#B^\pi$ explodes and computations become intractable like in standard probabilistic POMDPs. The next section presents the first contribution of this paper, which exploits a specific structure of the problem that is very common in practice.

## 3 Possibilistic and qualitative mixed observable MDPs ($\pi$-MOMDPs)

The complexity issue of $\pi$-POMDP solving is due to the fact that the size of the belief state space $B^\pi$ exponentially grows with the size of the state space $\mathcal{S}$. However, in practice, states are rarely totally hidden. Using mixed observability can be a solution: inspired by a similar recent work in probabilistic POMDPs [9, 1], we present in this section a structured modeling that takes into account situations where the agent directly observes some part of the state. This model generalizes both $\pi$-MDPs and $\pi$-POMDPs.

Like in [1], we assume that the state space $\mathcal{S}$ can be written as a Cartesian product of a visible state space $\mathcal{S}_v$ and a hidden one $\mathcal{S}_h$: $\mathcal{S} = \mathcal{S}_v \times \mathcal{S}_h$. Let $s = (s_v, s_h)$ be a state of the system. The component $s_v$ is directly observed by the agent and $s_h$ is only partially observed through the observations of the set $\mathcal{O}_h$: we denote by $\pi(o'_h \mid s', a)$ the possibility distribution over the future observation $o'_h \in \mathcal{O}_h$ knowing the future state $s' \in \mathcal{S}$ and the current action $a \in \mathcal{A}$. Figure 1 illustrates this structured model.

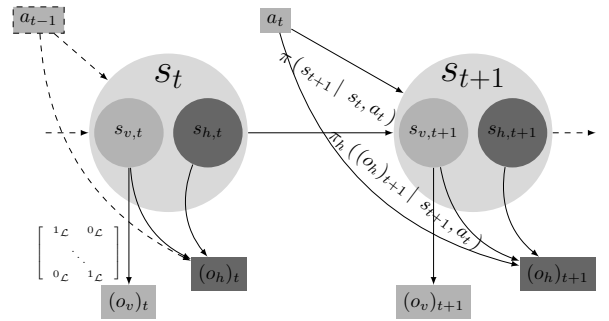

Figure 1: Graphical representation of a $\pi$-MOMDP

The visible state space is integrated to the observation space: $\mathcal{O}_v = \mathcal{S}_v$ and $\mathcal{O} = \mathcal{O}_v \times \mathcal{O}_h$. Then, knowing that the current visible component of the state is $s_v$, the agent *necessarily* observes $o_v = s_v$ (if $o'_v \neq s_v$, $\pi(o'_v \mid s_v) = 0_\mathcal{L}$). Formally, seen as a $\pi$-POMDP, its observation possibility distribution can be written as:

$$\begin{aligned}\pi(o' \mid s', a) &= \pi(o'_v, o'_h \mid s'_v, s'_h, a) \\ &= \min\{\pi(o'_h \mid s'_v, s'_h, a), \pi(o'_v \mid s'_v)\} \\ &= \begin{cases} \pi(o'_h \mid s', a) & \text{if } o'_v = s'_v \\ 0_\mathcal{L} & \text{otherwise} \end{cases}\end{aligned} \quad (4)$$

since $\pi(o'_v \mid s'_v) = 1_\mathcal{L}$ if $s'_v = o'_v$ and $0_\mathcal{L}$ otherwise. The following theorem, based on this equality enables the belief over hidden states to be defined.

**Theorem 1.** *Each reachable belief state of a $\pi$-MOMDP can be written as an element of $\mathcal{S}_v \times B_h^\pi$ where $B_h^\pi$ is the set of possibility distributions over $\mathcal{S}_h$: any $\beta \in B^\pi$ can be written as $(s_v, \beta_h)$ with $\beta_h(s_h) = \max_{\bar{s}_v \in \mathcal{S}_v} \beta(\bar{s}_v, s_h)$ and $s_v = \arg\max_{\bar{s}_v \in \mathcal{S}_v} \beta(\bar{s}_v, s_h)$.*

*Proof.* We proceed by induction on $t \in \mathbb{N}$: as the initial visible state $s_{v,0}$ is known by the agent, only states $s = (s_v, s_h)$ for which $s_v = s_{v,0}$ are such that $\beta_0(s) > 0_\mathcal{L}$. A belief over hidden states can be thus defined as $\beta_{h,0}(s_h) = \max_{s_v \in \mathcal{S}_v} \beta_0(s_v, s_h) = \beta_0(s_{v,0}, s_h)$.

At time $t$, if $\beta_t(s) = 0_\mathcal{L}$ for each $s = (s_v, s_h) \in \mathcal{S}$ such that $s_v \neq s_{v,t}$, the same notation can be adopted: $\beta_{h,t}(s_h) = \beta_t(s_{v,t}, s_h)$. Thus, if the agent reaches state $s_{t+1} = (s_{v,t+1}, s_{h,t+1})$ and if $s' = (s'_v, s'_h)$ with $s'_v \neq s_{v,t+1}$, then $s'_v \neq o_{v,t+1}$ and:

$$\begin{aligned}\pi(o_{t+1}, s' \mid \beta_t, a_t) &= \min\{\pi(o_{t+1} \mid s', a_t), \beta_{t+1}^{a_t}(s')\} \\ &= 0_\mathcal{L}.\end{aligned}$$

thanks to Equation (4). Finally, update Formula (3) ensures that $\beta_{t+1}(s') = 0_\mathcal{L}$. Then, $\beta_{t+1}$ is entirely encoded by $(s_{v,t+1}, \beta_{h,t+1})$ with $s_{v,t+1} = o_{v,t+1}$ and $\beta_{h,t+1}(s_h) = \max_{s_v} \beta_{t+1}(s_v, s_h) \ \forall s_h \in \mathcal{S}_h$. □

As all needed belief states are in $\mathcal{S}_v \times B_h^\pi$, the next theorem redefines the dynamic programming equation restricted to this product space.

**Theorem 2.** *Over $\mathcal{S}_v \times B_h^\pi$, the dynamic programming equation becomes:* $\forall i \in \{1, \ldots, p\}, \forall t \in \mathbb{N}$,
$u_i^*(s_v, \beta_h)$
$$= \max_{a \in \mathcal{A}} \max_{s'_v \in \mathcal{S}_v} \max_{o'_h \in \mathcal{O}_h} \min\left\{\beta^a(s'_v, o'_h), u_{i-1}^*(s'_v, \beta_h^{a, s'_v, o'_h})\right\}$$

*with the initialization* $u_0^*(s_v, \beta_h) = \mu(s_v, \beta_h)$,

*where* $\mu(s_v, \beta_h) = \min_{s_h \in \mathcal{S}_h} \max\{\mu(s_v, s_h), n(\beta_h(s_h))\}$
*is the preference over $\mathcal{S}_v \times B_h^\pi$,*

$$\beta^a(s'_v, o'_h) = \max_{s'_h \in \mathcal{S}_h} \min\{\pi(o'_h \mid s'_v, s'_h, a), \beta^a(s'_v, s'_h)\},$$

*and the belief update* $\beta_h^{s'_v, o'_h, a}(s'_h)$
$$= \begin{cases} 1_\mathcal{L} \text{ if } \min\{\pi(o'_h \mid s'_v, s'_h, a), \beta^a(s'_v, s'_h)\} \\ \qquad = \beta^a(s'_v, o'_h) > 0_\mathcal{L} \\ \min\{\pi(o'_h \mid s'_v, s'_h, a), \beta^a(s'_v, s'_h)\} \text{ otherwise} \end{cases}$$

*Proof.* Using the classical dynamic programming equation, Theorem 1, and the fact that $\mathcal{S}_v = \mathcal{O}_v$,

$$\begin{aligned}u_i^*(s_v, \beta_h) &= u_i^*(\beta) \\ &= \max_{a \in \mathcal{A}} \max_{(o'_v, o'_h) \in \mathcal{O}} \min\left\{\beta^a(o'_v, o'_h), u_{i-1}^*(\beta^{a,(o'_v, o'_h)})\right\} \\ &= \max_{a \in \mathcal{A}} \max_{s'_v \in \mathcal{S}_v} \max_{o'_h \in \mathcal{O}_h} \min\left\{\beta^a(s'_v, o'_h), u_{i-1}^*(\beta^{a, s'_v, o'_h})\right\} \\ &= \max_{a \in \mathcal{A}} \max_{s'_v \in \mathcal{S}_v} \max_{o'_h \in \mathcal{O}_h} \min\left\{\beta^a(s'_v, o'_h), u_{i-1}^*(s'_v, \beta_h^{a, s'_v, o'_h})\right\}\end{aligned}$$

where $\forall s_h \in \mathcal{S}_h$, $\beta_h^{a, s'_v, o'_h}(s_h) = \max_{\bar{s}_v} \beta^{a, s'_v, o'_h}(\bar{s}_v, s_h) = \beta^{a, s'_v, o'_h}(s'_v, s_h)$. For the initialization, we just note that $n(\beta(\bar{s}_v, s_h)) = 1_\mathcal{L}$ when $\bar{s}_v \neq s_v$, then

$$\begin{aligned}\mu(\beta) &= \min_s \max\{\mu(s), n(\beta(s))\} \\ &= \min_{s_h} \max\{\mu(s_v, s_h), n(\beta(s_h, s_v))\},\end{aligned}$$

which completes the procedure. The belief over observations defined in the last section can be written: $\forall o' = (o'_v, o'_h) \in \mathcal{O}$,

$$\begin{aligned}\beta^a(o') &= \max_{s' \in \mathcal{S}} \min\{\pi(o'_v, o'_h \mid s', a), \beta_t^{a_t}(s')\} \\ &= \max_{s'_h \in \mathcal{S}_h} \min\{\pi(o'_h \mid s'_v, s'_h, a_t), \beta_t^{a_t}(s'_v, s'_h)\}\end{aligned}$$

with $s'_v = o'_v$ since otherwise $\pi(o' \mid s'_v, s'_h, a) = 0_\mathcal{L}$ according to Equation (4). Then: $\beta^a(s'_v, o'_h) = \beta^a(o'_v, o'_h)$. Finally, using the standard update Equation (3) with $o'_v = s'_v$ and Equation (4), we get the new belief update. □

A standard algorithm would have computed $u_p^*(\beta)$ for each $\beta \in B^\pi$ while this new dynamic programming equation leads to an algorithm which computes it only for all $(s_v, \beta_h) \in \mathcal{S}_v \times B_h^\pi$. The size of the new belief space is $\#(\mathcal{S}_v \times B_h^\pi) = \#\mathcal{S}_v \times (\#\mathcal{L}^{\#\mathcal{S}_h} - (\#\mathcal{L} - 1)^{\#\mathcal{S}_h})$, which is exponentially smaller than the size of standard $\pi$-POMDPs' belief space: $\#\mathcal{L}^{\#\mathcal{S}_v \times \#\mathcal{S}_h} - (\#\mathcal{L} - 1)^{\#\mathcal{S}_v \times \#\mathcal{S}_h}$.

## 4 Solving $\pi$-MOMDPs

A finite policy for possibilistic MOMDPs can now be computed for larger problems using the dynamic programming equation of Theorem 2 and selecting maximizing actions for each state $(s_v, \beta_h) \in \mathcal{S}_v \times B^\pi$, as done in Equation (2) for each $s \in \mathcal{S}$. However, for many problems in practice, it is difficult to determine a horizon size. The goal of this section is to present an algorithm to solve $\pi$-MOMDPs with infinite horizon, which is the first proved algorithm to solve $\pi$-(MO)MDPs.

### 4.1 The $\pi$-MDP case

Previous work, [12, 14], on solving $\pi$-MDPs proposed a Value Iteration algorithm that was proved to compute optimal value functions, but not necessarily optimal policies for some problems with cycles. There is a similar issue in *undiscounted* probabilistic MDPs where the greedy policy at convergence of Value Iteration does not need to be optimal [11]. It is not surprising that we are facing the same issue in $\pi$-MDPs since the possibilistic dynamic programming operator does not rely on algebraic products so that it cannot be contracted by some *discount factor* $0 < \gamma < 1$.

---

**Algorithm 1:** $\pi$-MDP Value Iteration Algorithm

**for** $s \in \mathcal{S}$ **do**
  $u^*(s) \leftarrow 0_\mathcal{L}$ ;
  $u^c(s) \leftarrow \mu(s)$ ;
  $\delta(s) \leftarrow \overline{a}$ ;
**while** $u^* \neq u^c$ **do**
  $u^* = u^c$ ;
  **for** $s \in \mathcal{S}$ **do**
    $u^c(s) \leftarrow \max_{a \in \mathcal{A}} \max_{s' \in \mathcal{S}} \min\{\pi(s' \mid s,a), u^*(s')\}$ ;
    **if** $u^c(s) > u^*(s)$ **then**
      $\delta(s) \in \operatorname{argmax}_{a \in \mathcal{A}} \max_{s' \in \mathcal{S}} \min\{\pi(s' \mid s,a), u^*(s')\}$ ;

**return** $u^*$, $\delta$ ;

---

To the best of our knowledge, we propose here the first Value Iteration algorithm for $\pi$-MDPs, that provably returns an optimal policy, and that is different from the one of [14]. Indeed, in the deterministic example of Figure 2, action $\overline{a}$, which is clearly suboptimal, was found to be optimal in state $s_1$ with this algorithm: however it is clear that since $\pi(s_2 \mid s_1, b) = 1_\mathcal{L}$ and $\mu(s_2) = 1_\mathcal{L}$, $u_1^*(s_1) = 1_\mathcal{L}$. Obviously, $u_1^*(s_2) = 1_\mathcal{L}$ and since $\pi(s_1 \mid s_1, \overline{a}) = 1_\mathcal{L}$, $\max_{s' \in \mathcal{S}} \min\{\pi(s' \mid s_1, a), u_1^*(s')\} = 1_\mathcal{L}$ $\forall a \in \{\overline{a}, b\} = \mathcal{A}$, i.e. all actions are optimal in $s_1$. The "if" condition of Algorithm 1 permits to select the optimal action $b$ during the first step. This condition and the initialization, which were not present in previous algorithms of the literature, are needed to prove the optimality of the policy. The proof, which is quite lengthy and intricate, is presented in Appendix A. This sound algorithm for $\pi$-MDPs will then be extended to $\pi$-MOMDPs in the next section.

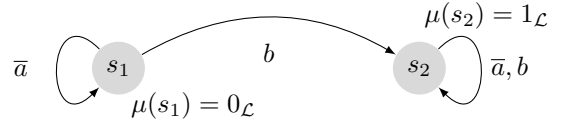

Figure 2: Example

As mentioned in [12], we assume the existence of an action "stay", denoted by $\overline{a}$, which lets the system in the same state with necessity $1_\mathcal{L}$. This action is the possibilistic counterpart of the discount parameter $\gamma$ in the probabilistic model, in order to guarantee convergence of the Value Iteration algorithm. However, we will see action $\overline{a}$ is finally used only on some particular satisfactory states. We denote by $\overline{\delta}$ is the decision rule such that $\forall s \in \mathcal{S}$, $\overline{\delta}(s) = \overline{a}$. The set of all the finite policies is $\Delta = \cup_{i \geqslant 1} \Delta_i$, and $\#\delta$ is the size of a policy $(\delta)$ in terms of decision epochs. We can now define the optimistic criterion for an infinite horizon: if $(\delta) \in \Delta$,

$$u(s, (\delta)) = \max_{\tau \in \mathcal{T}_{\#\delta}} \min\{\Pi(\tau \mid s, (\delta)), M(\tau)\}. \quad (5)$$

**Theorem 3.** *If there exists an action $\overline{a}$ such that, for each $s \in \mathcal{S}$, $\pi(s' \mid s, \overline{a}) = 1_\mathcal{L}$ if $s' = s$ and $0_\mathcal{L}$ otherwise, then Algorithm 1 computes the maximum optimistic criterion and an optimal policy which is stationary i.e. which does not depend on the stage of the process $t$.*

*Proof.* See Appendix A. □

Let $s$ be a state such that $\delta(s) = \overline{a}$, where $\delta$ is the returned policy. By looking at Algorithm 1, it can be noted that $u^*(s)$ always remains equal to $0_\mathcal{L}$ during the algorithm: $\forall s' \in \mathcal{S}$, either $\forall a \in \mathcal{A}$ $\mu(s) \geqslant \pi(s' \mid s, a)$, or $\mu(s) \geqslant u^*(s')$. If the problem is non trivial, it means that $s$ is a goal ($\mu(s) > 0_\mathcal{L}$) and that degrees of possibility of transitions to better goals are less than the degree of preference for $s$.

### 4.2 Value Iteration for $\pi$-MOMDPs

We are now ready to propose the Value Iteration algorithm for $\pi$-MOMDPs. In order to clarify this algorithm, we set

$$U(a, s'_v, o'_h, \beta_h) = \min\left\{\beta^a(s'_v, o'_h), u^*(s'_v, \beta_h^{a, s'_v, o'_h})\right\}.$$

Note that Algorithm 2 has the same structure as Algorithm 1. Note that a $\pi$-MOMDP is a $\pi$-MDP over $\mathcal{S}_v \times B_h^\pi$. Let $s_v \in \mathcal{S}_v$, $\beta_h \in B_h^\pi$ and now $\Gamma_{\beta, \overline{a}, s'_v}(\beta'_h) = \left\{o'_h \in \mathcal{O}_h \mid \beta_h^{\overline{a}, s'_v, o'_h} = \beta'_h\right\}$. To satisfy the assumption of Theorem 3, it suffices to ensure that $\max_{o'_h \in \Gamma_{\beta, \overline{a}, s'_v}(\beta'_h)} \beta^{\overline{a}}(s'_v, o'_h) = 1_\mathcal{L}$ if $s'_v = s_v$ and $\beta'_h = \beta_h$ and $0_\mathcal{L}$ otherwise. This property is verified when $\pi(s' \mid s, \overline{a}) = 1_\mathcal{L}$ if $s' = s$ (and $0_\mathcal{L}$ otherwise) and there exists an observation "nothing" $\overline{o}$ that is required for each state when $\overline{a}$ is chosen: $\pi(o' \mid s', \overline{a}) = 1_\mathcal{L}$ if $o' = \overline{o}$ and $0_\mathcal{L}$ otherwise.

**Algorithm 2:** $\pi$-MOMDP Value Iteration Algorithm

**for** $s_v \in \mathcal{S}_v$ and $\beta_h \in B_h^\pi$ **do**
    $u^*(s_v, \beta_h) \leftarrow 0_\mathcal{L}$ ;
    $u^c(s_v, \beta_h) \leftarrow \mu(s_v, \beta_h)$ ;
    $\delta(s_v, \beta_h) \leftarrow \overline{a}$ ;

**while** $u^* \neq u^c$ **do**
    $u^* = u^c$ ;
    **for** $s_v \in \mathcal{S}_v$ and $\beta_h \in B_h^\pi$ **do**
        $u^c(s) \leftarrow \max_{a \in \mathcal{A}} \max_{s'_v \in \mathcal{S}} \max_{o'_h \in \mathcal{O}_h} U(a, s'_v, o'_h, \beta_h)$ ;
        **if** $u^c(s_v, \beta_h) > u^*(s_v, \beta_h)$ **then**
            $\delta(s) \in \operatorname{argmax}_{a \in \mathcal{A}} \max_{s'_v \in \mathcal{S}} \max_{o'_h \in \mathcal{O}_h} U(a, s'_v, o'_h, \beta_h)$ ;

**return** $u^*$, $\delta$ ;

## 5 Experimental results

Consider a robot over a grid of size $g \times g$, with $g > 1$. It always perfectly knows its location on the grid $(x, y) \in \{1, \ldots, g\}^2$, which forms the visible state space $\mathcal{S}_v$. It starts at location $s_{v,0} = (1, 1)$. Two targets are located at $(x_1, y_1) = (1, g)$ ("target 1") and $(x_2, y_2) = (g, 1)$ ("target 2") on the grid, and the robot perfectly knows their positions. One of the targets is $A$, the other $B$ and the robot's mission is to identify and reach target $A$ as soon as possible. The robot does not know which target is $A$: the two situations, "target 1 is $A$" ($A1$) and "target 2 is $A$" ($A2$), constitute the hidden state space $\mathcal{S}_h$. The moves of the robot are deterministic and its actions $\mathcal{A}$ consist in moving in the four directions plus the action "stay".

At each stage of the process, the robot analyzes pictures of each target and gets then an observation of the targets' natures: the two targets ($oAA$) can be observed as A, or target 1 ($oAB$), or target 2 ($oBA$) or no target ($oBB$).

In the probabilistic framework, the probability of having a good observation of target $i \in \{1, 2\}$, is not really known but approximated by $Pr(good_i \mid x, y) = \frac{1}{2}\left[1 + \exp\left(-\frac{\sqrt{(x-x_i)^2 + (y-y_i)^2}}{D}\right)\right]$ where $(x, y) = s_v \in \{1, \ldots, g\}^2$ is the location of the robot, $(x_i, y_i)$ the position of target $i$, and $D$ a normalization constant. Then, for instance, $Pr(oAB \mid (x, y), A1)$ is equal to $Pr(good_1 \mid (x, y)) Pr(good_2 \mid (x, y))$, $Pr(oAA \mid (x, y), A1)$ to $Pr(good_1 \mid (x, y)) \times [1 - Pr(good_2 \mid (x, y))]$, and so on. Each step of the process before reaching a target costs 1, reaching target $A$ is rewarded by 100, and -100 for $B$. The probabilistic policy was computed in mixed-observability settings with APPL [9], using a precision of 0.046 (the memory limit is reached for higher precisions) and $\gamma = 0.99$. This problem can not be solved with the exact algorithm for MOMDPs [1] because it consumes the entire RAM after 15 iterations.

Using qualitative possibility theory, it is always possible to observe the good target: $\pi(good \mid x, y) = 1$. Here $\mathcal{L}$ will be a finite subset of $[0, 1]$, that is why $1_\mathcal{L}$ can be denoted by 1. Secondly, the more the robot is far away from target $i$, the more likely it can badly observe it (e.g. observe $A$ instead of $B$), which is a reasonable assumption concerning the imprecisely known observation model: $\pi(bad_i \mid x, y) = \frac{\sqrt{(x-x_i)^2 + (y-y_i)^2}}{\sqrt{2}(g-1)}$. Then for instance, $\pi(oAB \mid (x, y), A1) = 1$, $\pi(oAA \mid (x, y), A1) = \pi(bad_2 \mid x, y)$, $\pi(oBA \mid (x, y), A1) = \min\{\pi(bad_1 \mid x, y), \pi(bad_2 \mid x, y)\}$, etc. Note that the situation is fully known when the robot is at a target's location: thus there is no risk of being blocked in an unsatisfactory state, that is why using the *optimistic* $\pi$-MOMDP works. $\mathcal{L}$ thus consists in 0, 1, and all the other intermediate possible values of $\pi(bad \mid x, y)$. Note that the construction of this model with a probability-possibility transformation [4] would have been equivalent. The preference distribution $\mu$ is equal to 0 for all the system's states and to 1 for states $[(x_1, y_1), A1]$ and $[(x_2, y_2), A2]$ where $(x_i, y_i)$ is the position of target $i$. As mentioned in [12], the computed policy guarantees a shortest path to a goal state. The policy then aims at reducing the mission's time.

Standard $\pi$-POMDPs, which do not exploit mixed observability contrary to our $\pi$-MOMDP model, could not solve even very small $3 \times 3$ grids. Indeed, for this problem, $\#\mathcal{L} = 5$, $\#\mathcal{S}_v = 9$, and $\#\mathcal{S}_h = 2$. Thus, $\#\mathcal{S} = \#\mathcal{S}_v \times \#\mathcal{S}_h = 18$ and the number of belief states is then $\#B^\pi = \mathcal{L}^{\#\mathcal{S}} - (\mathcal{L}^{\#\mathcal{S}} - 1)^{\#\mathcal{S}} = 5^{18} - 4^{18} \geqslant 3.7 \cdot 10^{12}$ instead of 81 states with a $\pi$-MOMDP. Therefore, the following experimental results could **not** be conducted with standard $\pi$-POMDPs, which indeed justifies our present work on $\pi$-MOMDPs.

In order to compare performances of the probabilistic and possibilistic models, we compare *their total rewards at execution*: since the situation is fully known when the robot is at a target's location, it can not end up choosing target $B$. If $k$ is the number of time steps to identify and reach the correct target, then the total reward is $100 - k$.

We consider now that, in reality (thus here for the simulations), used image processing algorithms badly perform when the robot is far away from targets, *i.e.*, if $\forall i \in \{1, 2\}$, $\sqrt{(x-x_i)^2 + (y-y_i)^2} > C$, with $C$ a positive constant, $Pr(good_i \mid x, y) = 1 - P_{bad} < \frac{1}{2}$. In all other cases, we assume that the probabilistic model is the good one. We used $10^4$ simulations to compute the statistical mean of the total reward at execution.

The grid was $10 \times 10$, $D = 10$ and $C = 4$.

Figure 3.a shows that the probabilistic is more affected by the introduced error than the possibilistic one: it shows the total reward at execution of each model as a function of $P_{bad}$, the probability of badly observing tagets when the robot's location is such that $\sqrt{(x - x_i)^2 + (y - x_i)^2} > C$. This is due to the fact that the possibilistic update of the belief does not take into account new observations when the robot has already obtained a more reliable one, whereas the probabilistic model modifies the current belief at each step. Indeed, as there are only two hidden states (that we now denote by $s_h^1$ and $s_h^2$ ), if $\beta_h(s_h^1) < 1$, then $\beta_h(s_h^2) = 1$ (possibilistic normalization). The definition of the joint possibility of a state and an observation (minimum of the belief in state and observation possibilities) imply that the joint possibility of $s_h^1$ and the obtained observation, is smaller than $\beta_h(s_h^1)$. The possibilistic counterpart of the belief update equation (3) then ensures that the next belief is either more skeptic about $s_h^1$ if the observation is more reliable and confirms the prior belief ($\pi\left(o_h \mid s_v, s_h^1, a\right)$ is smaller than $\beta_h(s_h^1)$); or changes to the opposite belief if the observation is more reliable and contradicts the prior belief ($\pi\left(o_h \mid s_v, s_h^2, a\right)$ is smaller than both $\beta_h(s_h^1)$ and $\pi\left(o_h \mid s_v, s_h^1, a\right)$); or yet simply remains unchanged if the observation is not more informative than the current belief. The probabilistic belief update does not have these capabilities to directly change to the opposite belief and to disregard less reliable observations: the robot then proceed towards the wrong target because it is initially far away and thus badly observes targets. When it is close to this target, it gets good observations and gradually modifies its belief which becomes true enough to convince it to go towards the right target. However it has to cross a remote area away from targets: this yet gradually modifies its belief, which becomes wrong, and the robot finds itself in the same initial situation: it loses thus a lot of time to get out of this loop. We can observe that the total reward increases for high probabilities of misperceiving $P_{bad}$: this is because this high error leads the robot to reach the wrong target faster, thus to entirely know that the true target is the other one.

Now if we set $P_{bad} = 0.8$ and evaluate the total reward at execution for different wrong initial beliefs, we get Figure 3.b with the same parameters: we compare here the possibilistic model and the probabilistic one when the initial belief is strongly oriented towards the wrong hidden states (i.e. the agent strongly believes that target 1 is B whereas it is A in reality). Note that the possibilistic belief of the good target decreases when the necessity of the bad one increases. This figure shows that the possibilistic model yields higher rewards at execution if the initial belief is wrong and the observation function is imprecise.

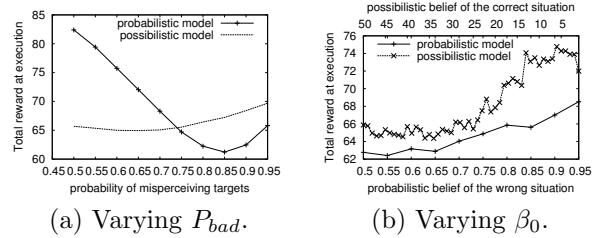

(a) Varying $P_{bad}$.    (b) Varying $\beta_0$.

Figure 3: Comparison of the total reward gathered at execution for possibilistic and probabilistic models.

## 6 Conclusion and perspectives

We have proposed a Value Iteration algorithm for possibilistic MDPs, which can produce optimal stationary policies in infinite horizon contrary to previous methods. We have provided a complete proof of convergence that relies on the existence of intermediate "stay" actions that vanish for non goal states in the final optimal policy. Finally, we have extended this algorithm to a new Mixed-Observable possibilistic MDP model, whose complexity is exponentially smaller than possibilistic POMDPs, so that we could compare $\pi$-MOMDPs with their probabilistic counterparts on realistic robotic problems. Our experimental results show that possibilistic policies can outperform probabilistic ones when the observation function yields imprecise results.

Qualitative possibilistic frameworks can however be inappropriate when some probabilistic information is actually available: POMDPs with Imprecise Parameters (POMDPIP) [7] and Bounded-parameter POMDPs (BPOMDP) [8] integrate the lack of knowledge by considering spaces of possible probability distributions. When such spaces can not be extracted or when a qualitative modeling suffices, $\pi$-POMDPs can be a good alternative, especially as POMDPIPs and BPOMDPs are extremely difficult to solve in practice. Yet, we plan to compare our $\pi$-MOMDP model with these imprecise probabilistic POMDPs in a near future.

The pessimistic version of $\pi$-MDPs can be easily constructed, but the optimality of the policy returned by the associated value iteration algorithm seems hard to prove, essentially because it is not enough to construct a maximizing trajectory, as the proof of section A does. The works [16, 10] could help us to get results about pessimistic $\pi$-MDP in order to solve unsafe problems.

## A Proof of Theorem 3

This appendix demonstrates that Algorithm 1 returns the maximum value of Equation (5) and an optimal policy. Note that the policy is optimal regardless of the initial state. We recall that $\exists \bar{a} \in \mathcal{A}$ such that $\forall s \in \mathcal{S}$, $\pi(s' \mid s, \bar{a}) = 1_\mathcal{L}$ if $s' = s$, and $0_\mathcal{L}$ otherwise. The existence of this action $\bar{a}$ makes the maximum value of the criterion non-decreasing with respect to the horizon size:

**Lemma 1.** $\forall s \in \mathcal{S}, \forall p \geqslant 0, u_p^*(s) \leqslant u_{p+1}^*(s).$

*Proof.* Let $s_0 \in \mathcal{S}$. $u_{p+1}^*(s_0)$
$$= \max_{\Delta_{p+1}} \max_{\tau \in \mathcal{T}_{p+1}} \min\left\{ \min_{i=0}^{p} \pi(s_{i+1} \mid s_i, \delta_i(s_i)), \mu(s_{p+1}) \right\}.$$
Consider the particular trajectories $\tau' \in \mathcal{T}'_{p+1} \subset \mathcal{T}_{p+1}$ such that $\tau' = (s_1, \ldots, s_p, s_p)$, and particular policies $(\delta') \in \Delta'_{p+1} \subset \Delta_{p+1}$ such that $(\delta') = (\delta_0, \ldots, \delta_{p-1}, \bar{\delta})$. It is obvious that $u_{p+1}^*(s_0) \geqslant$
$$\max_{(\delta') \in \Delta'_{p+1}} \max_{\tau' \in \mathcal{T}'_{p+1}} \min\left\{ \min_{i=0}^{p} \pi(s_{i+1} \mid s_i, \delta_i(s_i)), \mu(s_{p+1}) \right\}.$$
But note that the right part of this inequality can be rewritten as $\max_{(\delta) \in \Delta_p} \max_{\tau \in \mathcal{T}_p}$
$\min\left\{ \min_{i=0}^{p-1} \pi(s_{i+1} \mid s_i, \delta_i(s_i)), \pi(s_p \mid s_p, \bar{a}), \mu(s_p) \right\}$
$= u_p^*(s_0)$ since $\pi(s_p \mid s_p, \bar{a}) = 1_\mathcal{L}$. □

The meaning of this lemma is: it is always more possible to reach a state $s$ from $s_0$ in at most $p+1$ steps than in at most $p$ steps. As for each $s \in \mathcal{S}$, $(u_p^*(s))_{p \in \mathbb{N}} \leqslant 1_\mathcal{L}$, Lemma 1 insures that the sequence $(u_p^*(s))_{p \in \mathbb{N}}$ converges. The next lemma shows that the convergence of this sequence occurs in finite time.

**Lemma 2.** *For all $\forall s \in \mathcal{S}$, the number of iterations of the sequence $(u_p^*(s))_{p \in \mathbb{N}}$ up to convergence is bounded by $\#\mathcal{S} \times \#\mathcal{L}$.*

*Proof.* Recall first that values of the possibility and preference distributions are in $\mathcal{L}$ which is finite and totally ordered: we can write $\mathcal{L} = \{0_\mathcal{L}, l_1, l_2, \ldots, 1_\mathcal{L}\}$ with $0_\mathcal{L} < l_1 < l_2 < \ldots < 1_\mathcal{L}$. If two successive functions $u_k^*$ and $u_{k+1}^*$ are equal, then $\forall s \in \mathcal{S}$ sequences $(u^*(s))_{p \geqslant k}$ are constant. Indeed this sequence can be defined by the recursive formula
$$u_p^*(s) = \max_{a \in \mathcal{A}} \max_{s' \in \mathcal{S}} \min\left\{ \pi(s' \mid s, a), u_{p-1}^*(s') \right\}.$$
Thus if $\forall s \in \mathcal{S}, u_p^*(s) = u_{p-1}^*(s)$ then the next iteration $(p+1)$ faces the same situation $(u_{p+1}^*(s) = u_p^*(s) \ \forall s \in \mathcal{S})$. The slowest convergence can then be described as follows: for each $p \in \mathbb{N}$ only one $s \in \mathcal{S}$ is such that $u_{p+1}^*(s) > u_p^*(s)$. Moreover, for this $s$, if $u_p^*(s) = l_i$, then $u_{p+1}^*(s) = l_{i+1}$. We can conclude that for $p > \#\mathcal{L} \times \#\mathcal{S}$, the sequence is constant. □

First note that the variable $u^*(s)$ of Algorithm 1 is equal to $u_p^*(s)$ after the $p^{th}$ iteration. We conclude that $u^*$ converges to the maximal value of the criterion for an $(\#\mathcal{L} \times \#\mathcal{S})$-size horizon and can not be greater: the function $u^*$ returned is thus optimal with respect to Equation (5) and is computed in a finite number of steps.

In the following, we prove the optimality of the policy $(\delta^*)$ returned by Algorithm 1. For this purpose, we will construct a trajectory of size smaller than $\#\mathcal{S}$ which maximizes $\min\{\Pi(\tau \mid s_0, (\delta)), M(\tau)\}$ with policy $(\delta^*)$. The next two lemmas are needed for this construction and require some notations.

Let $s_0 \in \mathcal{S}$ and $p$ be the smallest integer such that $\forall p' \geqslant p$, $u_{p'}^*(s_0) = u^*(s_0)$, where $u^*$ is here the optimal value of the infinite horizon criterion of Equation (5) (variable $u^*(s)$ of Algorithm 1 does not increase after $p$ iterations). Equation (2) can be used to return an optimal policy (not stationary) denoted by $(\delta^{(s_0)}) \in \Delta_p$. With this notation: $\forall s \in \mathcal{S}, \ \delta^*(s) = \delta_0^{(s)}(s)$. Consider now a trajectory $\tau = (s_1, s_2, \ldots, s_p)$ which maximizes $\min\left\{ \min_{i=0}^{p-1} \pi\left(s_{i+1} \mid s_i, \delta_i^{(s_0)}(s_i)\right), \mu(s_p) \right\}$. This trajectory is called *optimal trajectory of minimal size from $s_0$*.

**Lemma 3.** *Let $\tau = (s_1, \ldots, s_p)$ be an optimal trajectory of minimal size from $s_0$.*
*Then, $\forall k \in \{1, \ldots, p-1\}, u^*(s_0) \leqslant u^*(s_k).$*

*Proof.* Let $k \in \{1, \ldots, p-1\}$.
$$u^*(s_0) = \min\left\{ \min_{i=0}^{p-1} \pi\left(s_{i+1} \mid s_i, \delta_i^{(s_0)}(s_i)\right), \mu(s_p) \right\}$$
$$\leqslant \min\left\{ \min_{i=k}^{p-1} \pi\left(s_{i+1} \mid s_i, \delta_i^{(s_0)}(s_i)\right), \mu(s_p) \right\}$$
$\leqslant u_{p-k}^*(s_k) \leqslant u^*(s_k)$ since $(u_p^*)_{p \in \mathbb{N}}$ is non-decreasing (Lemma 1). □

**Lemma 4.** *Let $\tau = (s_1, \ldots, s_p)$ be an optimal trajectory of minimal size from $s_0$ and $k \in \{1, \ldots, p-1\}$. If $u^*(s_0) = u^*(s_k)$, then $\delta^*(s_k) = \delta_k^{(s_0)}(s_k)$.*

*Proof.* Suppose that $u^*(s_0) = u^*(s_k)$. Since $u^*(s_0) \leqslant u_{p-k}^*(s_k) \leqslant u^*(s_k)$ (Lemma 3), we obtain that $u_{p-k}^*(s_k) = u^*(s_k)$. The criterion in $s_k$ is thus optimized within a $(p-k)$-horizon. Moreover a shorter horizon is not optimal: $\forall m \in \{1, \ldots, p-k\}$, $u_{p-k-m}^*(s_k) < u^*(s_k)$ i.e. with a $(p-k-m)$-size horizon the criterion in $s_k$ is not maximized. Indeed if the contrary was true, the criterion in $s_0$ would be maximized within a $(p-m)$-size horizon: the policy
$$\delta' = (\delta_0^{(s_0)}, \delta_1^{(s_0)}, \ldots, \delta_{k-1}^{(s_0)}, \delta_0^{(s_k)}, \ldots, \delta_{p-k-m-1}^{(s_k)}) \in \Delta_{p-m}$$
would be optimal. Indeed, $u^*(s_0)$
$$= \min\left\{ \min_{i=0}^{k-1} \pi\left(s_{i+1} \mid s_i, \delta_i^{(s_0)}(s_i)\right), u_{p-k}^*(s_k) \right\}$$
$$= \min\left\{ \min_{i=0}^{k-1} \pi\left(s_{i+1} \mid s_i, \delta_i^{(s_0)}(s_i)\right), u^*(s_k) \right\}$$

Then let $\bar{\tau} = (\bar{s}_1, \ldots, \bar{s}_{p-k-m}) \in \mathcal{T}_{p-k-m}$ be an optimal trajectory of minimal size from $s_k$. Setting $\bar{s}_0 = s_k$, $\bar{\tau}$ thus maximizes $u^*(s_k) = \min\left\{\min_{i=0}^{p-k-m-1} \pi\left(\bar{s}_{i+1} \mid \bar{s}_i, \delta_i^{(s_k)}(\bar{s}_i)\right), \mu(\bar{s}_{p-k-m})\right\}$.
If $(s'_1, \ldots, s'_{p-m}) = (s_1, \ldots, s_{k-1}, \bar{s}_0, \ldots, \bar{s}_{p-k-m})$, $u^*(s_0) = \min\left\{\min_{i=0}^{p-m-1} \pi\left(s'_{i+1} \mid s'_i, \delta'_i(s_i)\right), \mu(s'_{p-m})\right\}$ i.e. $\exists p' < p$ such that $u^*(s_0) = u^*_{p'}(s_0)$: it contradicts the assumption that $(s_1, \ldots, s_p)$ is an optimal trajectory of minimum size. Thus $p-k$ is the smallest integer such that $u^*_{p-k}(s_k) = u^*(s_k)$: we finally conclude that $\delta^*(s_k) (:= \delta_0^{(s_k)}(s_k)) = \delta_k^{(s_0)}(s_k)$. $\square$

**Theorem 4.** *Let $(\delta^*)$ be the policy returned by Algorithm 1; $\forall s_0 \in \mathcal{S}$, there exists $p^* \leqslant \#\mathcal{S}$ and a trajectory $(s_1, \ldots, s_{p^*})$ such that*

$$u^*(s_0) = \min\left\{\min_{i=0}^{p^*-1} \pi\left(s_{i+1} \mid s_i, \delta^*(s_i)\right), \mu(s_{p^*})\right\}:$$

*i.e. $\delta^*$ is an optimal policy.*

*Proof.* Let $s_0$ be in $\mathcal{S}$ and $\tau$ be an optimal trajectory of minimal size $p$ from $s_0$. If $\forall k \in \{1, \ldots, p-1\}$, $\delta^*(s_k) := \delta_0^{(s_k)}(s_k) = \delta_k^{(s_0)}(s_k)$ then the criterion in $s_0$ is maximized with $(\delta^*)$ since it is maximized with $(\delta^{(s_0)})$ and the optimality is shown. If not, let $k$ be the smallest integer $\in \{1, \ldots, p-1\}$ such that $\delta_0^{(s_k)}(s_k) \neq \delta_k^{(s_0)}(s_k)$. Lemmas 3 and 4 ensure that $u^*(s_k) > u^*(s_0)$. Definition of $k$ ensures that $u^*(s_k) > u^*(s_i) \ \forall i \in \{0, \ldots, k-1\}$.

Reiterate beginning with $s_0^{(1)} = s_k$: let $p^{(1)}$ be the number of iterations until variable $u^*(s^{(1)})$ of the algorithm converges (the smallest integer such that $u^*(s_0^{(1)}) = u^*_{p^{(1)}}(s_0^{(1)})$). Let $\tau^{(1)} \in T_{p^{(1)}}$ which maximizes $\min\{\min_{i=0}^{p^{(1)}-1} \pi(s_{i+1}|s_i, \delta_i^{(s_0^{(1)})}(s_i)), \mu(s_p^{(1)})\}$ ($\tau^{(1)}$ is an optimal trajectory of minimal size from $s_k = s_0^{(1)}$). We select $k^{(1)}$ in the same way as previously and reiterate beginning with $s_0^{(2)} = s_{k^{(1)}}^{(1)}$ which is such that $u^*(s_{k^{(1)}}^{(1)}) > u^*(s_0^{(1)})$, and $u^*(s_{k^{(1)}}^{(1)}) > u^*(s_i^{(1)}) \ \forall i \in \{0, \ldots, k^{(1)}-1\}$ etc... Lemma 5 below shows that all selected states $(s_0, \ldots, s_{k-1}, s_0^{(1)}, \ldots, s_{k^{(1)}-1}^{(1)}, s_0^{(2)}, \ldots, s_{k^{(2)}-1}^{(2)}, s_0^{(3)}, \ldots)$, are different. Thus this selection process ends since $\#\mathcal{S}$ is a finite set. The total number of selected states is denoted by $p^* = k + \sum_{i=1}^{q-1} k^{(i)} + p^{(q)}$ with $q \geqslant 0$ the number of new selected trajectories. Then the policy $(\delta') = (\delta_0, \ldots, \delta_{k-1}, \delta_0^{(s_0^{(1)})}, \ldots, \delta_{k^{(1)}-1}^{(s_0^{(1)})}, \ldots, \delta_{p^{(q)}}^{(s_0^{(q)})})$ corresponds to $(\delta^*)$ over $\tau' = (s'_1, \ldots, s'_{p^*}) = (s_0, s_1, \ldots, s_{k-1}, s_0^{(1)}, \ldots, s_{k^{(1)}-1}^{(1)}, \ldots, s_{p^{(q)}-1}^{(m)})$ and this policy is optimal because $u^*(s_0) = u(s_0, (\delta^*))$:

$$u^*(s_0) = \min\left\{\min_{i=0}^{k-1} \pi\left(s'_{i+1} \mid s'_i, \delta'(s'_i)\right), u^*_{p-k}(s_k)\right\}$$
$$\leqslant \min\left\{\min_{i=0}^{k-1} \pi\left(s'_{i+1} \mid s'_i, \delta'(s'_i)\right), u^*(s_k)\right\}$$
$$= \min\left\{\min_{i=0}^{k^{(1)}-1} \pi\left(s'_{i+1} \mid s'_i, \delta'(s'_i)\right), u^*_{p^{(1)}-k^{(1)}}(s_{k^{(1)}})\right\}$$
$$\ldots \leqslant \min\left\{\min_{i=0,\ldots,p^*-1} \pi\left(s'_{i+1} \mid s'_i, \delta'(s'_i)\right), \mu(s'_{p^*})\right\}$$

The "$\leqslant$" signs are in fact "$=$" since otherwise we would find a policy such that $u(s_0, (\delta')) > u^*(s_0)$. Thus $(\delta^*)$ is optimal: $u^*(s_0) = \min\left\{\min_{i=0}^{p^*-1} \pi\left(s'_{i+1} \mid s'_i, \delta^*(s'_i)\right), \mu(s'_{p^*})\right\}$ $\square$

**Lemma 5.** *The process described in the previous proof in order to construct a trajectory maximizing the criterion with $(\delta^*)$ always selects different system states.*

*Proof.* First, two equal states in the same selected trajectory $\tau^{(m)}$ would contradict the hypothesis that $p^{(m)}$ is the smallest integer such that $u^*_{p^{(m)}}(s_0^{(m)}) = u^*(s_0^{(m)})$. Indeed let $k$ and $l$ be such that $0 \leqslant k < l \leqslant p^{(m)}$ and suppose that $s_k^{(m)} = s_l^{(m)}$. For clarity in the next calculations, we omit "$(m)$": $p = p^{(m)}$ and $\forall i \in \{0, \ldots, l\}$, $s_i = s_i^{(m)}$. $u^*_{p-k}(s_k) = \min\{\min_{i=k}^{l-1} \pi\left(s_{i+1} \mid s_i, \delta_i^{(s_0)}(s_i)\right), u^*_{p-l}(s_l)\}$ $\leqslant u^*_{p-l}(s_l) = u^*_{p-l}(s_k)$. However $u^*_{p-k}(s_k) \geqslant u^*_{p-l}(s_k)$ (non-decreasing sequence).
We finally get $u^*_{p-k}(s_k) = u^*_{p-l}(s_k)$, thus

$$u^*(s_0) = \min\left\{\min_{i=0}^{k-1} \pi\left(s_{i+1} \mid s_i, \delta_i^{(s_0)}(s_i)\right), u^*_{p-k}(s_k)\right\}$$
$$= \min\left\{\min_{i=0}^{k-1} \pi\left(s_{i+1} \mid s_i, \delta_i^{(s_0)}(s_i)\right), u^*_{p-l}(s_l)\right\}$$
$$= \min\left\{\min_{i=0,\ldots,k-1,l,\ldots,p-1} \pi\left(s_{i+1} \mid s_i, \delta_i^{(s_0)}(s_i)\right), \mu(s_p)\right\}$$

Consequently, a $(p^{(m)} - l + k)$-sized horizon is good enough to reach the optimal value: it is a contradiction. Finally, if we suppose that a state $\bar{s}$ appears two times in the sequence of selected states, then this state belongs to two different selected trajectories $\tau^{(m)}$ and $\tau^{(m')}$ (with $m' < m$). Lemma 3 and the definition of $k^{(m')}$ which implies that $u^*(s_0^{(m'+1)})$ is strictly greater than the criterion's optimal values in each of the states $s_0^{(m')}, \ldots, s_{k^{(m')}-1}^{(m')}$ requires that $u^*(s_0^{(m)}) \leqslant u^*(\bar{s}) < u^*(s_0^{(m'+1)})$. It is a contradiction because $u^*(s_0^{(m'+1)}) \leqslant u^*(s_0^{(m)})$ since $m' < m$. $\square$